\documentclass{article}

     \PassOptionsToPackage{numbers, compress}{natbib}

     \usepackage[final]{neurips_2024}

\usepackage[utf8]{inputenc} 
\usepackage[T1]{fontenc}    
\usepackage{hyperref}       
\usepackage{url}            
\usepackage{booktabs}       
\usepackage{amsfonts}       
\usepackage{nicefrac}       
\usepackage{microtype}      
\usepackage{xcolor}         
\usepackage{amsmath}        
\usepackage{graphicx}
\usepackage{multirow}
\usepackage{makecell}

\title{Slide2Text: Leveraging LLMs for Personalized Textbook Generation from PowerPoint Presentations}
\author{
    Yizhou Zhou$^{1, 2, 3}$\thanks{Corresponding Author: zhouyizhou25@stu.ecnu.edu.cn}, 
    \\
    $^{1}$~Lab for Artificial Intelligence for Education, East China Normal University, Shanghai, China \\
    $^{2}$~Shanghai Institute of Artificial Intelligence for Education, East China Normal University, Shanghai, China \\
    $^{3}$~School of Computer Science and Technology, East China Normal University, Shanghai, China  \\
}

\begin{document}
\maketitle
\begin{abstract}
The rapid advancements in Large Language Models (LLMs) have revolutionized educational technology, enabling innovative approaches to automated and personalized content creation. This paper introduces Slide2Text, a system that leverages LLMs to transform PowerPoint presentations into customized textbooks. By extracting slide content using OCR, organizing it into a coherent structure, and generating tailored materials such as explanations, exercises, and references, Slide2Text streamlines the textbook creation process. Flexible customization options further enhance its adaptability to diverse educational needs. The system highlights the potential of LLMs in modernizing textbook creation and improving educational accessibility. Future developments will explore multimedia inputs and advanced user customization features.
\end{abstract}

\section{Introduction}
The rapid evolution of artificial intelligence, particularly Large Language Models (LLMs), has profoundly reshaped educational technology\cite{doughty_comparative_2024}. Their advanced natural language processing capabilities, including high-quality text generation, semantic understanding, and personalized content recommendations, provide innovative solutions to longstanding challenges in educational content creation, such as inefficiency, lack of personalization, and limited scalability\cite{liu_chatqa_2024}.

Personalized instruction and efficient content generation have become essential in contemporary educational environments. However, traditional methods of textbook creation remain labor-intensive, time-consuming, and inadequately tailored to the diverse needs of learners. PowerPoint (PPT) presentations, widely used for instructional delivery, contain rich but often fragmented textual and visual content, making their systematic and coherent conversion into structured textbooks technically challenging, particularly when extracting textual information embedded within images and ensuring content accuracy.

Recent advancements in generative AI have introduced promising opportunities to automate educational content generation. Large Language Models (LLMs), when guided by structured prompts, can effectively convert PPT slides into logically structured textbooks. These models offer significant customization capabilities, enabling users to define language styles, organize chapters, integrate authoritative references, and generate exercises. Nevertheless, critical challenges such as accurate text extraction from slides (particularly images), managing external knowledge integration, and ensuring content reliability have remained. The integration of Retrieval-Augmented Generation (RAG) techniques represents a significant advancement, effectively addressing content accuracy and enriching textbook materials with authoritative knowledge sources.

To effectively address these challenges, this paper introduces Slide2Text V3, an innovative system designed to automate the transformation of PPT slides into personalized, high-quality textbooks. Utilizing a modular and extensible architecture enhanced with RAG functionality, Slide2Text streamlines the entire content generation workflow—from precise slide content extraction and structured textbook generation to customized educational outputs enriched with validated external references. Additionally, Slide2Text leverages GitHub for collaborative development and efficient version management, enabling seamless system updates, scalability, and community-driven enhancements.

\section{Methodology and System Implementation}

\subsection{System Overview}
The Slide2Text system automates the transformation of PowerPoint presentations into structured and customized textbooks through a modular and extensible architecture. The latest version (V3) introduces advanced functionalities, including Retrieval-Augmented Generation (RAG) for enhanced content accuracy and GitHub-based collaborative management for efficient version control. The conversion process comprises four primary stages: content extraction, chapter structure generation, chapter content generation, and final content assembly. Intermediate data is serialized into JSON format, facilitating easy data exchange, flexibility, and future scalability. By systematically organizing multimodal PPT content, Slide2Text guarantees logical coherence, adaptive customization, and maintains high content fidelity.

During the content extraction phase, the system utilizes the \texttt{python-pptx} library to systematically parse textual content, layout information, and embedded metadata from presentation slides. Concurrently, to ensure comprehensive content capture, especially from graphical elements, the \texttt{pytesseract} OCR library is integrated to accurately extract textual information from images. The extracted data—including textual descriptions, image references, and metadata such as slide titles and presenter notes—is structured into a standardized JSON format. This structured format greatly simplifies subsequent content processing, retrieval, and integration into the textbook generation workflow.

In the textbook generation phase, Slide2Text leverages Large Language Models (LLMs) and Retrieval-Augmented Generation (RAG) techniques to deliver high-quality, tailored educational content through three core functions:

\begin{enumerate}
\item \textbf{PPT Information Extraction}: This function processes PowerPoint presentations to extract structured information, such as text, headings, and visual elements. By converting slide content into a machine-readable format, it lays the foundation for subsequent content generation, ensuring fidelity to the original material.

\item \textbf{RAG-Enhanced Retrieval}: Utilizing RAG techniques, this function augments content generation by retrieving authoritative knowledge from external academic databases and online resources. It ensures that the generated textbook content is comprehensive, accurate, and aligned with scholarly standards, bridging gaps between slide material and broader educational context.

\item \textbf{Personalized LLM Generation}: This function employs LLMs to generate textbook content tailored to user-specific needs. By incorporating customization parameters—such as language style, difficulty level, and educational objectives—it produces highly adaptable outputs, enabling educators to create textbooks that meet diverse learning requirements.
\end{enumerate}

Slide2Text's modular architecture ensures flexibility and scalability, supporting seamless integration with collaborative development platforms like GitHub and enabling future enhancements through community contributions. The system is poised for cloud-based deployment with technologies such as Docker and Kubernetes, enhancing cross-platform compatibility and scalability, while planned advancements in multimedia processing and user interface design aim to broaden its educational impact. As a transformative tool in automated content creation, Slide2Text empowers educators to efficiently produce high-quality, personalized textbooks, addressing varied pedagogical needs with precision and innovation.

\subsection{PPT Content Extraction and Textbook Structure Design}

\subsubsection{Text and Structure Extraction}

The Slide2Text system employs the Python-PPTX library to systematically parse and extract textual and structural content from PowerPoint files. During this extraction process, the system meticulously preserves the hierarchical organization of slide content, effectively capturing slide titles, main text bodies, bullet points, and associated speaker notes. Each slide within the PowerPoint presentation is sequentially traversed through the Python-PPTX presentation.slides object, ensuring accurate extraction and mapping of each text element to maintain the original logical order and knowledge framework presented in the source slides. This structured extraction is crucial as it forms the foundational data layer for subsequent processing, analysis, and content generation within the Slide2Text workflow.

\subsubsection{Image Processing and OCR}

To accurately interpret and incorporate textual information embedded within slide images, the system integrates the Tesseract Optical Character Recognition (OCR) engine. This process involves multiple carefully designed image preprocessing steps to enhance OCR accuracy, including image contrast enhancement, noise reduction, grayscale conversion, and resolution adjustments. The Slide2Text OCR module robustly handles complex textual scenarios, supporting multilingual recognition to accommodate mixed-language content, such as combined English and Chinese text. During processing, extracted slide images are temporarily saved as individual image files, passed to the Tesseract OCR engine, and the resulting textual outputs are systematically associated with their respective image sources for clear traceability.

\subsubsection{Data Integration and Storage}

All textual and visual content extracted from the PPT files, along with associated metadata, is systematically organized and persistently stored using the JSON (JavaScript Object Notation) format. Each slide’s data is structured into clearly defined JSON objects containing the following primary fields:

\begin{itemize}
\item \texttt{raw\_text}: Text extracted from slide titles, bodies, bullet points, and notes.
\item \texttt{ocr\_text}: Text derived from OCR processing of slide images.
\item \texttt{images}: Array of metadata for processed images, including file paths and OCR text.
\end{itemize}

This standardized, structured JSON storage approach ensures consistent compatibility across the Slide2Text system, providing a robust foundation for accurate content retrieval, analysis, textbook generation, and subsequent user-specific customization processes.

\subsection{LLM-based Content Generation}

\subsubsection{Multi-model Integration Framework}

Slide2Text implements a flexible multi-model integration framework that enables users to customize the content generation process by selecting from a variety of Large Language Model (LLM) APIs, such as OpenAI's GPT-4o or DeepSeek V3. The system provides pre-configured API interfaces for each supported model, allowing users to specify their preferred model based on task requirements or performance preferences. This user-driven selection ensures adaptability to diverse educational needs while maintaining seamless integration with the chosen LLM, supporting high-quality textbook generation tailored to specific pedagogical goals.

\subsubsection{Prompt Engineering}

To ensure the relevance, accuracy, and educational value of generated content, Slide2Text employs advanced prompt engineering strategies tailored for textbook creation. Given the constraints of content length and the potential decline in generation quality for overly long outputs, the system first performs chapter-wise segmentation, generating each chapter independently to maintain consistency and quality. For each chapter, the prompt is carefully constructed by incorporating multiple components: the corresponding content extracted from PowerPoint slides, user-defined customization requirements, and supplementary information retrieved through Retrieval-Augmented Generation (RAG) techniques. Additionally, the prompt leverages structured expressions to define the LLM's role as an "expert textbook author," alongside detailed generation requirements and constraints, such as maintaining scholarly rigor, ensuring pedagogical clarity, and adhering to specific structural formats. This comprehensive prompt design ensures that the generated content aligns with both the original slide material and the user’s educational objectives, delivering coherent and high-quality textbook chapters.

\subsection{RAG from Papers and Google}

\subsubsection{PDF Processing and Knowledge Base Construction}

Slide2Text harnesses an advanced Retrieval-Augmented Generation (RAG) framework to enhance the depth, accuracy, and scholarly authority of generated educational content. The process begins with extracting knowledge from academic PDFs using PyMuPDF, which efficiently captures textual content while preserving critical insights from scholarly publications relevant to the targeted educational domain. To ensure semantic coherence, the extracted text is segmented into smaller chunks using LangChain's RecursiveCharacterTextSplitter. Each chunk is set to approximately 200 characters, with a slight overlap to maintain contextual continuity, and segmentation prioritizes natural boundaries such as sentence endings to avoid breaking meaningful units. These segments are then converted into Markdown format, ensuring structured organization and seamless integration into the subsequent content generation pipeline.

\subsubsection{Vector Database and Retrieval}

The processed knowledge segments are encoded into high-dimensional vector representations using OpenAI’s \texttt{text-embedding-ada-002} model, which generates 1536-dimensional embeddings for precise semantic matching. These embeddings are stored in a FAISS (Facebook AI Similarity Search) database, utilizing the \texttt{IndexFlatL2} indexing structure to enable rapid and accurate retrieval. To optimize retrieval efficiency, the system adopts indexing configurations recommended by large language models, balancing speed and accuracy for large-scale knowledge bases. This setup ensures persistent storage and efficient updates, providing a robust foundation for retrieving relevant knowledge during content generation.

\subsubsection{Hybrid Reference Retrieval and Integration}

Slide2Text employs a hybrid retrieval strategy that combines local knowledge base queries with real-time external searches via the Google Custom Search API. The FAISS database enables semantic retrieval of pre-processed academic resources, ensuring access to highly relevant local references. Concurrently, the system retrieves up-to-date online content through Google’s API, enriching static knowledge with dynamic, current information. A weighted scoring algorithm integrates these local and external results, prioritizing references that are both authoritative and contextually aligned with the educational content being generated.

\subsubsection{Reference Generation and Quality Control}

To uphold academic integrity, Slide2Text systematically formats citations with a focus on usability and trustworthiness. Academic references are enriched with a relevance score, calculated based on semantic alignment with the generated content, addressing a common pain point in RAG systems by providing readers with clear indicators of reference reliability. For web-based references, the system includes concise content snippets and direct URLs, styled similarly to Google search results, to offer intuitive and accessible citations. Quality control is further enhanced through semantic similarity checks to eliminate duplicates, ensuring that only credible, high-value references are included in the final textbook.

\begin{figure}
    \centering
    \includegraphics[width=0.75\linewidth]{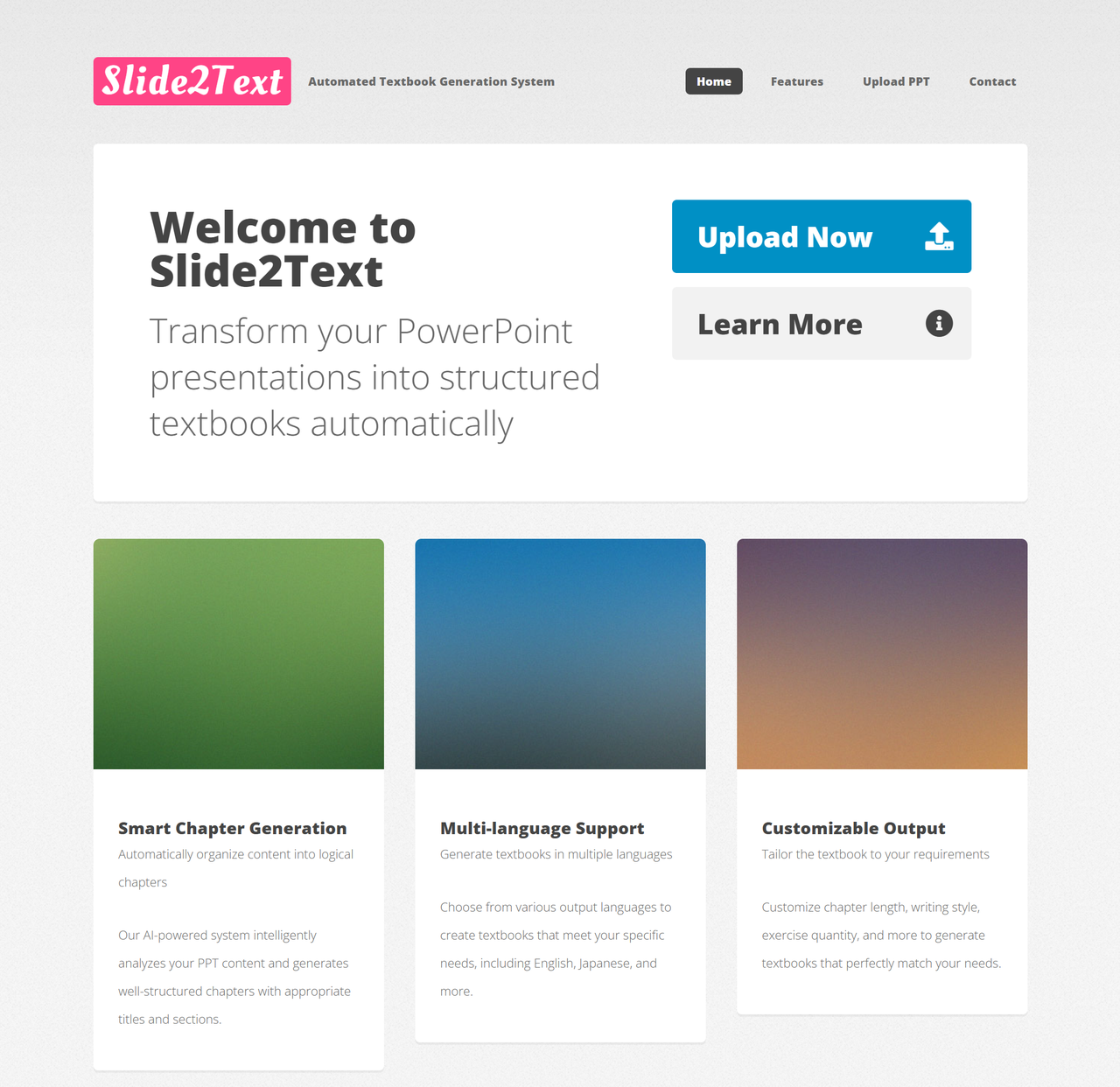}
    \caption{Websit of Slide2Text}
    \label{fig:enter-label}
\end{figure}

\subsection{Front-end Web Design and Links}

\subsubsection{User Interface Design}

Slide2Text’s user interface is built upon the Flask web framework, adopting modern responsive design principles to ensure usability across various devices, including desktops, tablets, and smartphones. Core features of the web interface include an intuitive file upload interface, enabling users to easily upload PowerPoint files through drag-and-drop or multi-file selection capabilities. The interface also provides real-time progress feedback with clear, visually informative indicators, enhancing user engagement and trust. Additionally, an interactive result preview allows users to view generated textbooks directly within the browser and download them conveniently, facilitating rapid review and iterative customization.

\subsubsection{System Integration and Performance Optimization}

Slide2Text ensures seamless functionality through a well-architected backend system that integrates robust routing, secure file handling, and efficient API interactions. Comprehensive URL routes manage main page interactions, file uploads, content processing triggers, and results display, delivering a coherent user experience. The system implements strict validation mechanisms to support secure uploading of PPT and PPTX files, minimizing potential security vulnerabilities, while a RESTful API infrastructure enables asynchronous processing and supports external integrations, laying the groundwork for future scalability.

To enhance performance and responsiveness, Slide2Text incorporates several optimization strategies within its architecture. Asynchronous operations are employed for file uploads and content generation, allowing efficient parallel task execution and improving the overall user experience. The system leverages advanced caching with in-memory stores like Redis to cache intermediate data and final outputs, significantly reducing redundant computations and accelerating response times. Additionally, automated routines manage temporary file storage by periodically cleaning up unused resources, ensuring optimal system performance and resource efficiency. Together, these strategies create a user-friendly, scalable, and high-performing platform for educational content generation, suitable for diverse academic and professional applications.

\section{Experiments and Results}

\subsection{Case Study: DX from PPT to Textbook Generation}

To evaluate the Slide2Text system in a real-world educational context, we conducted a case study using a PowerPoint presentation titled \textit{Technical Design Concept for Digital Transformation (DX)}, comprising 39 slides on diverse technical topics related to digital transformation, including textual explanations, diagrams, images, and speaker notes. The study was carried out in an English-taught course at a Japanese university, where both instructors and students required flexible language options to accommodate their linguistic preferences. Slide2Text enabled the generation of textbooks in both English and Japanese, reflecting its capability for personalized customization to meet the needs of diverse educational settings.

The evaluation simulated an authentic educational scenario, converting the PPT materials into structured, high-quality textbooks through several key stages. Initially, the system extracted content from the slides, including text, structural elements, and OCR-processed image data, preserving the original logical hierarchy in a structured JSON format. Subsequently, leveraging Large Language Models (LLMs) enhanced by Retrieval-Augmented Generation (RAG), Slide2Text generated a coherent chapter structure, complete with chapter titles, summaries, and authoritative external references. The final textbooks were produced in English and Japanese, each tailored with user-defined styles (academic or simplified), and included theoretical explanations, real-world case studies, interactive exercises, and validated references. This multilingual support and customization underscored Slide2Text’s effectiveness in addressing diverse pedagogical needs, facilitating seamless adoption by both instructors and students in the Japanese university setting.

\subsection{Experimental Evaluation Framework}

To comprehensively assess Slide2Text’s performance, quality, and educational impact, we designed a two-tiered evaluation framework comprising a baseline layer for quality assurance and an experimental layer for pedagogical optimization. This framework ensures that generated textbooks meet essential standards before deployment and validates their effectiveness in enhancing learning outcomes.

\subsubsection{Baseline Layer: Quality Assurance through Human-AI Collaboration}

The baseline layer focuses on a preliminary screening process to ensure that LLM-generated textbooks meet minimum standards in content quality, cultural inclusivity, and interactivity, safeguarding their suitability for educational use. This process employs a human-AI collaborative approach, combining automated evaluation with expert review. Initially, automated assessments are conducted using multiple LLMs (DeepSeekR1, ChatGPT-4o, and Gemini 2.0 Flash) through pairwise comparisons, scoring, and boolean question answering. Custom Python scripts further evaluate factual accuracy, formatting consistency, keyword coverage, and potential inappropriate content, leveraging LLMs’ efficiency in handling repetitive tasks. A report of identified issues—such as factual errors, cultural biases, or lack of engagement—is generated for review. Subsequently, an expert panel of educators and education technologists conducts a thorough evaluation, focusing on academic rigor, cultural sensitivity, and pedagogical appropriateness. The panel holds final authority on approving textbooks, ensuring they meet predefined quality thresholds in content accuracy, inclusivity, and interactivity before proceeding to experimental deployment.

\subsubsection{Experimental Layer: Pedagogical Optimization through Personalized Learning}

The experimental layer validates the pedagogical effectiveness of LLM-generated personalized textbooks, guided by personalized learning theory, using undergraduate students at Kyushu University enrolled in the \textit{Digital Transformation} course. The core research question is whether LLM-generated personalized textbooks, informed by pre-course learner analysis, outperform traditional materials in enhancing learning outcomes. The experiment employs a pre-test/post-test control group design, with students randomly assigned to an experimental group (using personalized textbooks) or a control group (using traditional PDF materials). Personalized textbooks are generated based on pre-course data, analyzing knowledge gaps, learning styles, and motivation levels to tailor content difficulty, presentation, and tone, while both groups’ materials share the same core knowledge derived from course PPTs and academic resources.

Participants are recruited voluntarily with informed consent, targeting at least 30 students per group to account for statistical power and attrition. Pre- and post-test questionnaires, including the AMS-C 28 motivation scale, assess knowledge mastery, learning motivation, and user experience, while open-ended feedback provides qualitative insights. The experimental procedure spans preparation (learner analysis and textbook generation), implementation (course delivery with personalized or traditional materials), and evaluation (post-test and feedback collection). Data analysis includes independent samples t-tests to compare post-test scores and motivation changes, paired t-tests for within-group improvements, and thematic analysis of qualitative feedback to identify strengths, limitations, and optimization opportunities for LLM-generated personalized textbooks.

\section{Conclusion and Future Work}

\subsection{Conclusion}

This study presented Slide2Text, a sophisticated AI-powered system designed to automate the conversion of PowerPoint presentations into customized, high-quality textbooks. Through modular integration of advanced LLMs, OCR technologies, and Retrieval-Augmented Generation (RAG), Slide2Text significantly streamlines educational content production. The system's multilingual capabilities, extensive customization options, and intuitive Flask-based web interface provide educators with powerful tools to rapidly develop personalized educational materials.

Case studies demonstrated Slide2Text’s ability to generate coherent, comprehensive textbooks, validated through rigorous automated and expert-led assessments. Despite its robust performance, ongoing challenges related to occasional LLM inaccuracies and limited interactive learning features remain priorities for future development.

\subsection{Future Work}

To address existing limitations and enhance Slide2Text’s educational impact, future efforts will target the following strategic areas:

\begin{enumerate} \item \textbf{Integration of Interactive Learning Elements}: Future iterations will incorporate interactive exercises, automated assessments, personalized learner feedback, and adaptive content recommendations based on student engagement metrics, fostering deeper learning interactions and improved educational outcomes.

\item \textbf{Enhanced Content Credibility and Verification}:
Strengthening the current RAG integration by incorporating robust verification mechanisms linked to scholarly databases (Semantic Scholar, Google Scholar) and authoritative knowledge bases (Wikipedia, discipline-specific repositories). This will minimize inaccuracies and improve the academic reliability of generated content through structured citations and reference sections.

\item \textbf{Advanced Logical Coherence and Multi-turn Contextual Generation}:
Refining prompt engineering strategies and chapter segmentation methods to effectively handle complex or interdisciplinary content. Leveraging advanced multi-turn contextual capabilities of newer LLMs, the system will achieve superior narrative coherence across textbook chapters and thematic structures.

\item \textbf{Expansion of Domain-specific Knowledge Graph Integration}:
Integrating structured, domain-specific knowledge graphs will provide a robust semantic foundation for textbook content, facilitating the accurate representation of complex knowledge structures and enhancing educational quality and depth.

\item \textbf{Sophisticated Human-AI Evaluation Collaboration}:
Continuously refining the human-AI collaborative evaluation framework, emphasizing pairwise comparisons, automated scoring, and Boolean question-answering methodologies. Employing multiple LLM-based evaluators will further improve the objectivity, reliability, and efficiency of textbook assessments, reducing evaluator biases and workload.

\end{enumerate}

Through these targeted enhancements, Slide2Text aspires to evolve into a comprehensive educational platform, combining automated efficiency with human expertise, ensuring high-quality, reliable, and interactive educational content that meets diverse learner needs in the digital era.

\section*{Acknowledgements}

This work was partially supported by the National Natural Science Foundation of China under Grant 61977058, and the Natural Science Foundation of Shanghai under Grant 23ZR1418500.

\bibliographystyle{IEEEtran}
\bibliography{reference}

\end{document}